\documentclass[10pt,twocolumn,letterpaper]{article}

\usepackage{cvpr}              
\definecolor{cvprblue}{rgb}{0.21,0.49,0.74}
\usepackage[pagebackref,breaklinks,colorlinks,allcolors=cvprblue]{hyperref}

\def\paperID{4944} 
\def\confName{CVPR}
\def\confYear{2026}

\usepackage{amsmath,amsfonts,bm}

\def\eqref#1{equation~\ref{#1}}

\def\1{\bm{1}}

\DeclareMathAlphabet{\mathsfit}{\encodingdefault}{\sfdefault}{m}{sl}
\SetMathAlphabet{\mathsfit}{bold}{\encodingdefault}{\sfdefault}{bx}{n}
\newcommand{\tens}[1]{\bm{\mathsfit{#1}}}

\def\tH{{\tens{H}}}

\def\tW{{\tens{W}}}

\usepackage{hyperref}
\usepackage{url}
\usepackage{booktabs}
\usepackage{multirow}
\usepackage{array}
\usepackage{listings}
\usepackage{xcolor}
\usepackage{graphicx}
\usepackage[most]{tcolorbox}
\usepackage{pgf-pie}
\usepackage{tikz}
\usetikzlibrary{arrows.meta}
\usetikzlibrary{shapes.geometric, positioning}
\usepackage{subcaption}
\usepackage{float}
\usepackage{adjustbox}
\usepackage{graphicx}
\usepackage{stfloats} 
\usepackage{makecell}
\usepackage[section]{placeins} 
\usepackage{tabularx}

\usepackage{pgfplots}
\usepgfplotslibrary{statistics}
\usepgfplotslibrary{polar}
\pgfplotsset{compat=1.18}
\usepackage{caption}
\newcommand{\nobrk}[1]{\mbox{#1}}

\usepackage{tikz}
\usepackage{amsmath}
\usepackage{xcolor}
\usepackage{graphicx}
\usetikzlibrary{calc,positioning,fit,arrows.meta}

\title{The Spatial Blindspot of Vision-Language Models}

\author{
Nahid Alam$^{1,\dagger}$, 
Leema Krishna Murali$^{1,5,\dagger}$,
Siddhant Bharadwaj$^{2}$,
Patrick Liu$^{3,\dagger}$,
Timothy Chung$^{4,1}$,
\and
Drishti Sharma$^{1}$,
Akshata A.$^{1,\dagger}$,
Kranthi Kiran$^{1,6}$, 
Wesley Tam$^{6,\dagger}$,
Bala Krishna S Vegesna$^7$,
\and 
\\
$^1$Cohere Labs Community, 
$^2$Indian Institute of Science, Bangalore,
$^3$UIUC, \\
$^4$Imperial College London, 
$^5$Eisai Inc., 
$^6$EleutherAI,
$^7$Georgia Institute of Technology\\
\\
$\dagger$~Work done as part of the EleutherAI SOAR Program\\
{\tt\small {nahid.m.alam@gmail.com}}
}

\begin{document}
\maketitle
\begin{abstract}
Vision-language models (VLMs) have advanced rapidly, but their ability to capture spatial relationships remains a blindspot. Current VLMs are typically built with contrastive language-image pretraining (CLIP) style image encoders. The training recipe often flattens images into 1D patch sequences, discarding the 2D structure necessary for spatial reasoning. We argue that this lack of spatial awareness is a missing dimension in VLM design and a bottleneck for applications requiring spatial grounding, such as robotics and embodied AI. To address this, we investigate (i) image encoders trained with alternative objectives and (ii) 2D positional encodings. Our experiments show that these architectural choices can lead to improved spatial reasoning on several benchmarks. 
\end{abstract}    
\section{Introduction}
\label{sec:intro}

Vision–language models (VLMs) have achieved impressive results in recognition, captioning, and visual question answering, yet their ability to reason about space remains fragile. Although spatial structure is fundamental to visual perception, most VLMs treat images primarily as sources of semantic tokens, rather than as organized physical layouts. This limits their reliability in tasks requiring geometric reasoning, precise localization, and spatial interaction.

A source of this weakness lies in the visual foundations of modern VLMs. They rely on large, pre-trained image encoders \citep{dosovitskiy2021imageworth16x16words, radford2021learning, sun2023evaclipimprovedtrainingtechniques, oquab2024dinov2learningrobustvisual}, typically frozen or lightly adapted \citep{alayrac2022flamingo, liu2023llava, li2023blip, jeong2024visualstylepromptingswapping}, that are optimized for global semantic alignment. Prior work shows that CLIP-style models emphasize overall semantics over local structure \citep{tong2024eyeswideshutexploring} and exhibit poor robustness to simple geometric transformations \citep{anis2025limitationsvisionlanguagemodelsunderstanding}. While recent efforts explore alternative pretraining objectives \citep{tong2024eyeswideshutexploring, maninis2025tipstextimagepretrainingspatial, tschannen2025siglip2multilingualvisionlanguage, fini2024multimodalautoregressivepretraininglarge}, their impact on spatial reasoning within VLMs remains underexplored.

Spatial information is further degraded during vision-language alignment. Most existing VLMs flatten visual tokens into a 1D sequence and apply RoPE-1D \citep{su2023roformerenhancedtransformerrotary}, a design choice that has been shown to weaken spatial awareness \citep{zhang2025scalingbeyondadvancingspatial, zhang2025mllmsstrugglespatialunderstanding}. Qwen2-VL \citep{wang2024qwen2} shows that explicitly factorizing positional encoding along spatial axes can improve structural preservation. Their findings suggest that alignment design is an area worth exploring for spatial understanding. 

Foundational VLMs \citep{alayrac2022flamingo, liu2023llava, li2023blip, xiao2024florence} are rarely evaluated on benchmarks that directly probe spatial reasoning. Standard evaluations emphasize semantic competence, offering limited insight into models’ understanding of geometry, relative position, and the scene layout.

In this work, we treat spatial grounding as a first-class design axis in VLMs and make the following contributions:

\begin{enumerate}
	\item We evaluate state-of-the-art VLMs on spatial reasoning benchmarks and identify gaps between semantic competence and spatial understanding.
	\item We study alternative image encoders—SigLIP, SigLIP2, and AIMv2—within the LLaVA framework and quantify how encoder choice affects spatial reasoning.
    \item We use 2D-RoPE for vision-language alignment to preserve the image’s 2D structure and improve spatial reasoning performance on several benchmarks.
\end{enumerate}

\section{Related Work}
\label{sec:relatedwork}
\subsection{Image Encoders}

In recent years, we have seen progress on image encoders aligned with their own text captioning. CLIP \citep{radford2021learning} aligns image and text representations through a large-scale contrastive loss on paired dataset, and demonstrates strong zero-shot generalization across classification and retrieval tasks. In contrast, DINOv2 \citep{oquab2024dinov2learningrobustvisual} is a visual model trained with a self-supervised masked image modeling objective. This allows DINOv2 to learn rich, pixel-level features by predicting masked-out image patches, without relying on paired text supervision. BLIP \citep{li2022blipbootstrappinglanguageimagepretraining} proposed a unified framework that integrates both contrastive pre-training for vision and generative objectives such as image–text matching and captioning. BLIP demonstrated that combining discriminative and generative signals leads to richer cross-modal representations and improved downstream transfer. BLIP-2 \citep{li2023blip} further advanced this line by introducing a lightweight querying transformer that bridges frozen image encoders with large language models, enabling efficient vision–language alignment without end-to-end pretraining of massive multimodal models. SigLIP \citep{zhai2023sigmoid} improves CLIP by replacing the softmax contrastive loss with a pairwise sigmoid loss, removing batch-size dependencies and resulting in more stable training dynamics and efficient scaling. SigLIP2 \citep{tschannen2025siglip2multilingualvisionlanguage} extends this line of work by augmenting the SigLIP training recipe with self-distillation and masked prediction objectives, resulting in higher-quality dense features, and improved localization across vision–language benchmarks. SigLIP2 also introduces the NaFlex variant, which preserves native aspect ratios and supports variable sequence lengths. The NaFlex variant provides empirical gains on aspect-sensitive tasks such as OCR and document understanding. In contrast to these contrastive and bridging approaches, AIMv2 \citep{fini2024multimodalautoregressivepretraininglarge} introduces an autoregressive multimodal pretraining framework that unifies vision and language by pairing a vision encoder with a decoder trained to jointly generate image patches and text tokens. Cocchi et al. \citep{cocchi2025llavamorecomparativestudyllms} 
conduct a study that systematically pairs various LLMs with different visual backbones such as CLIP, DINOv2, SigLIP, and SigLIP2 to understand the strengths and limitations of various VLM integration strategies. Our work differs from theirs as we focus on spatial awareness along with the integration of M-RoPE in image encoders; specifically 2D-RoPE for images.

\subsection{Spatial Reasoning}
Zhang et al. \citep{zhang2025scalingbeyondadvancingspatial} classified spatial reasoning into two categories, static vs. dynamic. Static spatial reasoning refers to understanding spatial relationships in fixed configurations, where objects do not change positions over time. These tasks test whether a model can identify, compare, or infer spatial arrangements from still images or single states of a scene. For example, determining whether 'the book is to the left of the laptop' or 'the chair is behind the table' requires recognizing and reasoning about stable relative positions. Static reasoning often emphasizes relational concepts such as distance, orientation, containment, or perspective, without involving temporal changes. Dynamic spatial reasoning, on the other hand, involves scenarios in which the spatial configuration evolves. Here, models must account for motion, transformation, or sequential changes in a scene. Typical examples include predicting the outcome of an object’s movement, tracking shifting perspectives, or reasoning about cause-and-effect actions in space. Unlike static reasoning, dynamic tasks demand temporal and often causal understanding, requiring models to build an implicit representation of change and transformation. Another way to categorize spatial reasoning is 2D vs. 3D images. 2D reasoning focuses on relationships within 2D images, such as relative position, alignment, or counting, whereas 3D reasoning requires understanding depth, perspective, and volumetric relations. In this paper, we focus on static spatial awareness in 2D images. 

\subsection{VLMs and Spatial Awareness}
Recent advances in LLMs and general purpose image encoders, such as CLIP \citep{radford2021learning} and SigLIP \citep{zhai2023sigmoid} have significantly improved VLM capabilities. Architectures such as Flamingo \citep{alayrac2022flamingo}, LLaVA \citep{liu2023llava, liu2023improvedllava}, KOSMOS \citep{kosmos-g, peng2023kosmos}, Florence-2 \citep{xiao2024florence}, and Molmo \citep{deitke2024molmo} demonstrate strong performance in tasks such as image captioning, visual question answering (VQA) and complex reasoning. Qwen2-VL \citep{wang2024qwen2} introduced multimodal rotary position encoding (M-RoPE) and dynamic resolution techniques, while PaLI’s joint modality scaling and cross-lingual learning \citep{chen2022pali, chen2023pali} have improved vision-language understanding. These VLMs tend to rely on implicit visual features learned during training rather than explicitly structured representations of spatial information needed for world understanding. RoboSpatial \citep{song2025robospatialteachingspatialunderstanding} is an annotated dataset specifically designed for spatial understanding in VLMs that includes 1 million images, 5,000 3D scans, and 3 million annotated spatial relationships. The pairing of 2D egocentric images with 3D scans in RoboSpatial makes it suitable for both 2D and 3D spatial understanding tasks. SpatialVLM \citep{chen2024spatialvlmendowingvisionlanguagemodels} focuses on training VLMs with spatial reasoning dataset on the internet to improve their 3D spatial understanding. This framework enhances the ability of VLMs to recognize quantitative relationships between physical objects, such as distances and size differences, which are crucial for many real-world tasks. SpatialVLM has also shown potential in robotics as a tool for providing fine-grained reward annotations. MM-Spatial \citep{daxberger2025mmspatialexploring3dspatial} introduces CA-VQA, a new supervised fine-tuning dataset and benchmark focused on indoor scenes. It targets tasks such as spatial relationship reasoning, metric estimation, and 3D grounding. By training MM-Spatial on CA-VQA, the model achieves state-of-the-art performance in 3D spatial understanding, highlighting the value of incorporating depth and multiview cues. SpatialRGPT \citep{cheng2024spatialrgptgroundedspatialreasoning} introduces a data curation pipeline that enables effective learning of regional representations from 3D scene graphs. It also features a flexible plugin module for integrating depth information into the visual encoder of existing VLMs. This allows SpatialRGPT to accurately perceive the relative directions and distances of user-specified regions within a visual scene.

\subsection{Measuring Spatial Awareness}

There are a number of benchmarks evaluating VLMs on spatial awareness; spanning across tasks such as 2D layout understanding, 3D geometric reasoning, counting, and relational question answering. Multimodal Visual Pattern (MMVP) benchmark \citep{tong2024eyeswideshutexploring} is a set of 300 images in 9 categories, designed by observing CLIP-blind pairs.  CV-Bench \citep{tong2024cambrian1} includes both the 2D and 3D tasks around spatial relationships, object counting, depth ordering, and relative distance estimation. GQA \citep{hudson2019gqa} leverages scene graph annotations and functional programs to generate compositional questions, and includes subsets like compare, verify, and logical that specifically test spatial relations such as left, right, on, and under. Visual Spatial Reasoning (VSR) \citep{liu2023vsr} assesses fine-grained relational understanding through 66 annotated spatial relations in natural images, while TopViewRS \citep{li2024topviewrs} measures reasoning over semantic top-view maps. TallyQA \citep{acharya2018tallyqa} and CountBenchQA \citep{paliGemma2024} address object numerosity under occlusion and clutter.

\section{Experimental Setup}
\label{sec:experiment}

\subsection{Methods}
Our experiment is based on the LLaVA \citep{liu2023llava} framework as shown in Figure \ref{fig:llavaarch}. To explore spatial awareness in VLMs, we extend the LLaVA architecture by integrating alternative image encoders beyond CLIP along with 2D-RoPE to preserve the inherent 2D spatial structure as shown in Figure \ref{fig:pretrain_strategies}. Specifically, we experiment with CLIP, SigLIP, SigLIP2, AIMv2 along with a modified design to include 2D-RoPE. Our hypothesis is that these encoder designs will better capture hierarchical spatial cues and contextual relationships across image regions. Unlike standard RoPE, 2D-RoPE encodes both horizontal and vertical patch indices through concatenated sinusoidal embeddings, providing explicit spatial priors for attention. In our implementation, 2D-RoPE is applied after patch embeddings are projected into query and key vectors, ensuring that spatial relations are injected directly into the attention mechanism rather than at the raw patch level. We hypothesize that these design choices will build a VLM with better relative position understanding in 2D images.

\subsection{Training Strategies}
We train our models in two stages: pretraining and instruction fine-tuning. The training process uses the same dataset used in LLaVA pretraining and instruction tuning respectively. For each image $X_v$, we used the single-turn conversation data $$(X^1_q,X^1_a, \dots, X^T_q, X^T_a)$$ where $T$ is the total number of turns from LLaVA.

We conduct our experiments using a cluster of 8 NVIDIA A40 GPUs, each with 48 GB of VRAM. For both pretraining and fine-tuning, we resize input images to a resolution of 256×256 pixels.

The pretraining process focuses solely on training the projection matrix. We use a per-device batch size of 32, which results in a global batch size of 256. We optimize the model using the AdamW optimizer with a learning rate of 1e-3, and a cosine learning rate scheduler is applied. For fine-tuning, we perform a full model update. The fine-tuning configuration uses a learning rate of 2e-5, per-device batch size of 16, resulting in a global batch size of 128. 
\section{Results}
\label{sec:results}

\input{sec/viz}

We compare frontier models and LLaVA variants on various benchmarks. Our work is based on the LLaVA-1.5 7B model. Therefore, all encoders and their 2D-RoPE variants that we trained are 7B models. Frontier models between the 2B to 8B parameter range were compared with the LLaVA variants we trained. For comparison purposes, we used LLaVA-NeXT 7B \citep{liu2024llavanext}, LLaVA-OneVision-qwen2-7B-ov-hf \citep{li2024llavaonevisioneasyvisualtask}, Qwen2.5-VL-8B \citep{bai2025qwen25vltechnicalreport}, SmolVLM2-2.2B-Instruct \citep{marafioti2025smolvlm}, Gemma3-4b-it \citep{gemmateam2025gemma3technicalreport}, PaliGemma2-3b-mix-448 \citep{paliGemma2024} and Molmo-7B-D-0924 \citep{deitke2024molmo}.

\begin{table*}[t]
    \centering
    \caption{Comparison of multimodal models across MMMU\_Val, MME, CCBench, and SEEDBench-IMG. Values \underline{underlined} indicate the best-performing model overall, while values in \textbf{bold} highlight the best-performing LLaVA encoder variant in each benchmark.}
    \label{tab:vlm-benchmarks}
    \begin{tabular}{lcccc}
        \toprule
        Models & MMMU\_Val & MME & CCBench & \makecell{SEEDBench-IMG} \\
        \toprule
        Qwen2.5-VL & \underline{0.580} & \underline{0.826} & \underline{0.592} & \underline{0.770} \\
        LLaVA-NeXT & 0.376 & 0.632 & 0.243 & 0.696 \\
        LLaVA-OneVision & 0.479 & 0.712 & 0.549 & 0.767 \\
        SmolVLM2 & 0.416 & 0.640 & 0.231 & 0.713 \\
        Gemma3-4b-it & 0.473 & 0.620 & 0.369 & 0.655 \\
        PaliGemma & 0.307 & 0.580 & 0.333 & 0.715 \\
        Molmo & 0.491 & 0.662 & 0.367 & 0.746 \\
        LLaVA-1.5 & 0.322 & 0.589 & 0.084 & 0.601 \\
        \midrule
        LLaVA-2D-RoPE & 0.298 & 0.582 & \textbf{0.086} & 0.585 \\
        LLaVA-SigLIP & 0.303 & 0.559 & 0.071 & 0.567 \\
        LLaVA-SigLIP-2D-RoPE & \textbf{0.334} & 0.525 & 0.080 & 0.545 \\
        LLaVA-SigLIP2 & 0.309 & 0.495 & \textbf{0.086} & 0.548 \\
        LLaVA-SigLIP2-2D-RoPE & 0.323 & 0.542 & \textbf{0.086} & 0.527 \\
        LLaVA-AIMv2 & 0.314 & \textbf{0.573} & \textbf{0.086} & \textbf{0.595} \\
        LLaVA-AIMv2-2D-RoPE & 0.311 & 0.563 & 0.114 & 0.586 \\
        \bottomrule
    \end{tabular}
\end{table*}

\begin{table*}[t]
    \centering
    \caption{Comparison of frontier models and LLaVA variants across spatial understanding benchmarks. Values \underline{underlined} indicate the best-performing frontier model; values in \textbf{bold} indicate the best-performing LLaVA variant.}
    {\setlength{\tabcolsep}{4pt} 
    \begin{tabular}{lccccccc}
        \toprule
        Models & MMVP & \makecell{CV-Bench \\ 2D Overall} & TallyQA & \makecell{GQA \\ Overall} & VSR & \makecell{Top-\\ViewRS} & \makecell{Count-\\BenchQA} \\
        \toprule
        LLaVA-NeXT & 0.667 & 0.606 & 0.733 & \underline{63.786} & 63.994 & 0.409 & 0.515 \\
        \nobrk{LLaVA-OneVision} & 0.767 & 0.730 & 0.797 & 62.140 & 77.741 & 0.414 & 0.823 \\
        Qwen2.5-VL & \underline{0.770} & \underline{0.754} & 0.800 & 60.391 & \underline{89.116} & \underline{0.456} & \underline{0.891} \\
        SmolVLM2 & 0.687 & 0.577 & 0.729 & 50.574 & 71.277 & 0.416 & 0.692 \\
        \nobrk{Gemma3-4b-it} & 0.708 & 0.659 & 0.525 & 31.277 & 55.074 & 0.334 & 0.713 \\
        PaliGemma & 0.667 & 0.624 & 0.794 & 62.570 & 65.139 & 0.322 & 0.674 \\
        Molmo & 0.753 & 0.728 & \underline{0.808} & 55.295 & 76.432 & 0.323 & 0.858 \\
        LLaVA v1.5 & 0.577 & 0.490 & 0.707 & 33.225 & 55.810 & 0.384 & 0.468 \\
        \midrule
        \nobrk{LLaVA-2D-RoPE} & 0.513 & 0.443 & 0.654 & 34.433 & 57.201 & 0.283 & 0.290 \\
        LLaVA-SigLIP & 0.433 & 0.412 & 0.672 & 25.648 & 54.910 & 0.349 & 0.581 \\
        \nobrk{LLaVA-SigLIP-2D-RoPE} & 0.507 & 0.425 & 0.616 & \textbf{38.448} & 57.692 & 0.295 & 0.483 \\
        LLaVA-SigLIP2 & 0.427 & 0.442 & 0.684 & 23.970 & 52.701 & \textbf{0.371} & 0.532 \\
        \nobrk{LLaVA-SigLIP2-2D-RoPE} & 0.480 & 0.415 & 0.646 & 34.560 & 56.465 & 0.330 & 0.402 \\
        LLaVA-AIMv2 & 0.513 & \textbf{0.466} & \textbf{0.710} & 32.541 & 56.219 & 0.339 & \textbf{0.739} \\
        \nobrk{LLaVA-AIMv2-2D-RoPE} & \textbf{0.560} & 0.432 & 0.690 & 32.342 & \textbf{60.311} & 0.338 & 0.719 \\
        \bottomrule
    \end{tabular}
    }
    \label{tab:vlm-spatial--benchmarks}
\end{table*}

\subsection{Evaluating on General Purpose Benchmarks}
As shown in Table~\ref{tab:vlm-benchmarks}, Qwen2.5-VL achieves the strongest overall performance, obtaining the highest scores on MMMU\_Val, MME, CCBench, and SEEDBench-IMG. Among the LLaVA variants, AIMv2 demonstrates the most consistent results, particularly on MME and SEEDBench-IMG, although its 2D-RoPE counterpart provides only marginal gains on certain metrics and about 1.75\% drop in MME. The effect of 2D-RoPE is mixed across variants. For example, LLaVA-SigLIP improves about 10\% on MMMU\_Val, while other models show little benefit or even reduced performance. LLaVA-SigLIP-2D-RoPE achieves the strongest MMMU\_Val score among LLaVA encoders, but AIMv2 remains more competitive across multiple benchmarks. 

\subsection{Evaluating on Spatial Benchmarks}
\begin{figure*}[t]
  \centering
  \begin{minipage}[t]{0.2\textwidth}
    \centering
    \captionof{figure}{Example image from LLaVA-Bench (In-the-Wild) \citep{liu2023llava}.}
    \label{fig:asianfood_vqa}
    \vspace{0.4\baselineskip}
    \fbox{\includegraphics[width=\linewidth]{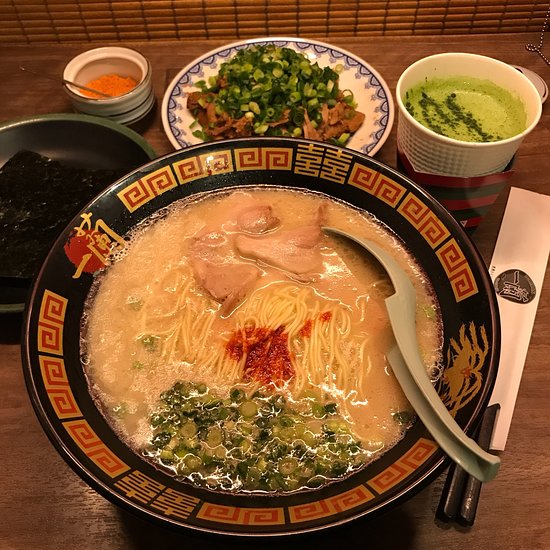}}
  \end{minipage}
  \hfill
  \begin{minipage}[t]{0.77\textwidth}
    \centering
    \captionof{table}{Model outputs for the prompt \textit{Are the chopsticks to the left or right of the bowl?} on image shown in Figure~\ref{fig:asianfood_vqa}.}
    \label{tab:asianfood_vqa_results}
    \begin{tabular}{l l}
        \toprule
        \textbf{Model} & \textbf{Output} \\
        \midrule
        LLaVA-v1.5 & The chopsticks are to the right of the bowl. \\
        LLaVA-2D-RoPE & The chopsticks are to the right of the bowl. \\
        LLaVA-SigLIP & The chopsticks are to the right of the bowl. \\
        LLaVA-SigLIP-2D-RoPE & The chopsticks are to the right of the bowl. \\
        LLaVA-SigLIP2 & The chopsticks are to the right of the bowl. \\
        LLaVA-SigLIP2-2D-RoPE & Right \\
        LLaVA-AIMv2 & The chopsticks are to the right of the bowl. \\
        LLaVA-AIMv2-2D-RoPE & The chopsticks are to the right of the bowl. \\
        Qwen2.5-VL & The chopsticks are to the right of the bowl. \\
        \textbf{Gemma3-4b-it} & \textbf{The chopsticks are to the left of the bowl.} \\
        \bottomrule
    \end{tabular}
  \end{minipage}
\end{figure*}

In Table~\ref{tab:vlm-spatial--benchmarks}, Qwen2.5-VL stands out as the strongest frontier model overall, achieving the highest performance on CV-Bench 2D, MMVP, VSR, TopViewRS, and CountBenchQA. LLaVA-NeXT leads frontier models on GQA Overall, while Molmo achieves the best TallyQA score among frontiers, indicating complementary strengths across different tasks. Among the LLaVA variants, performance is more fragmented: LLaVA-AIMv2 shows the most consistent improvements, reaching the highest scores on CV-Bench 2D, TallyQA, and CountBenchQA. LLaVA-AIMv2-2D-RoPE improves MMVP and leads on VSR. LLaVA-SigLIP-2D-RoPE dominates GQA Overall, and LLaVA-SigLIP2 leads in TopViewRS. We observe that although the LLaVA variants surpass Gemma3-4b-it in TallyQA, VSR; their overall performance did not surpass other frontier models. We think this is because of how Gemma3-4b-it is designed. Gemma3-4b-it uses SigLIP variant as its vision encoder, but the fixed resolution of the encoder with Pan \& Scan algorithm. Pan \& Scan is a preprocessing algorithm that allows the model to handle high-resolution and non-square images by breaking them down into smaller, fixed-size crops. This is necessary because Gemma 3's vision encoder can only process images at a fixed resolution of 896x896 pixels. This design leads to information loss that hinders performance in benchmarks that require fine-grained visual reasoning, such as VSR and TallyQA. In Table~\ref{tab:asianfood_vqa_results}, we observe that Gemma3-4b-it mistakenly thinks that the chopsticks are on the left side of the ramen bowl - reconfirming our hypothesis.

In Table~\ref{tab:vlm-spatial--benchmarks}, we also observe improved performance using AIMv2 encoder and its 2D-RoPE version over other vision encoders in the LLaVA framework. AIMv2 design focuses on the image-first principle, where the model is trained to process all the image patches before decoding the text tokens in an autoregressive manner. This dense per-token supervision helps with tasks requiring fine-grained perception, such as in MMVP, CV-Bench, TallyQA, CountBenchQA, and VSR. We also believe that the two-stage captioning pipeline to generate AIMv2 training data helps in count-related examples, causing the LLaVA-AIMv2 variant to perform well in TallyQA and CountBenchQA compared to other spatial benchmarks \citep{lai2024revisitlargescaleimagecaptiondata}.



\begin{figure}[H]
  \centering
  \begin{subfigure}[t]{0.45\linewidth}
    \centering
    \begin{subfigure}{0.45\linewidth}
      \centering
      \fbox{\includegraphics[width=\linewidth,height=\linewidth]{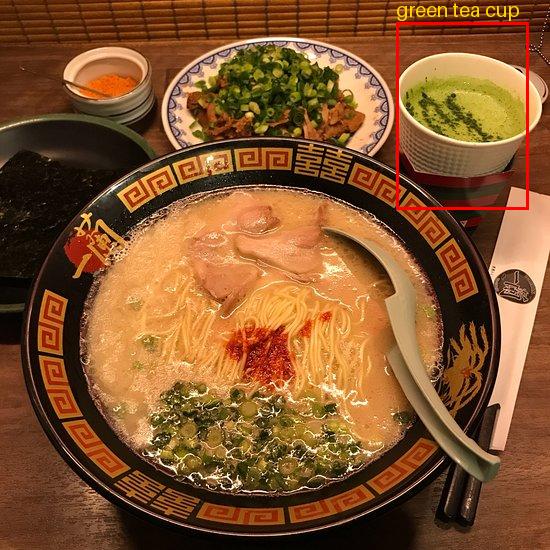}}
      \caption{SigLIP2}
      \label{fig:suba}
    \end{subfigure}
    \hfill
    \begin{subfigure}{0.45\linewidth}
      \centering
      \fbox{\includegraphics[width=\linewidth,height=\linewidth]{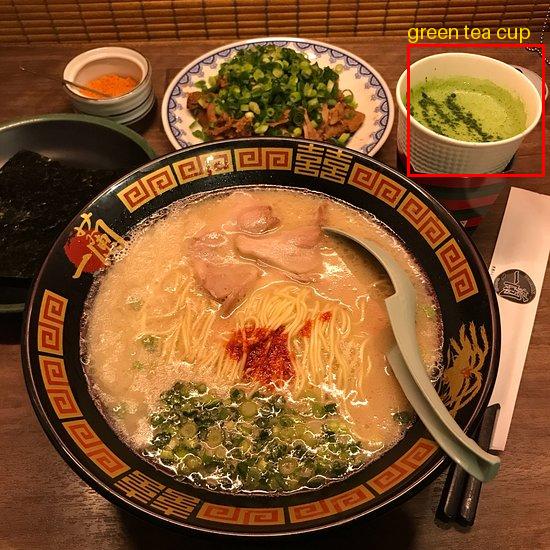}}
      \caption{AIMv2}
      \label{fig:subb}
    \end{subfigure}
    \caption{Output for prompt: \textit{Locate the cup that contains green liquid. Provide the bounding boxes.}}
    \label{fig:group1}
  \end{subfigure}
  \hfill
  \begin{subfigure}[t]{0.45\linewidth}
    \centering
    \begin{subfigure}{0.45\linewidth}
      \centering
      \fbox{\includegraphics[width=\linewidth,height=\linewidth]{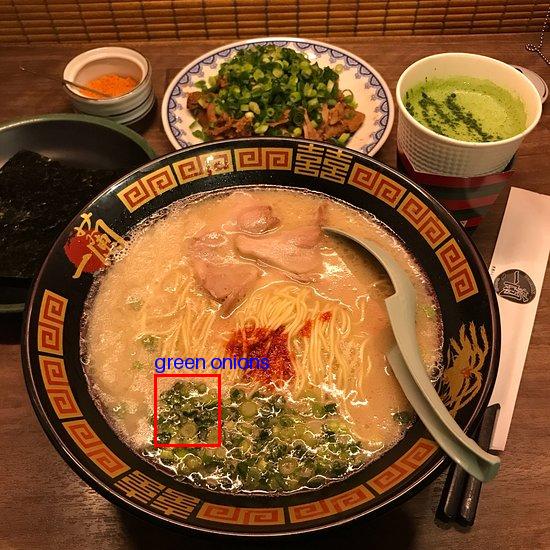}}
      \caption{SigLIP2}
      \label{fig:subc}
    \end{subfigure}
    \hfill
    \begin{subfigure}{0.45\linewidth}
      \centering
      \fbox{\includegraphics[width=\linewidth,height=\linewidth]{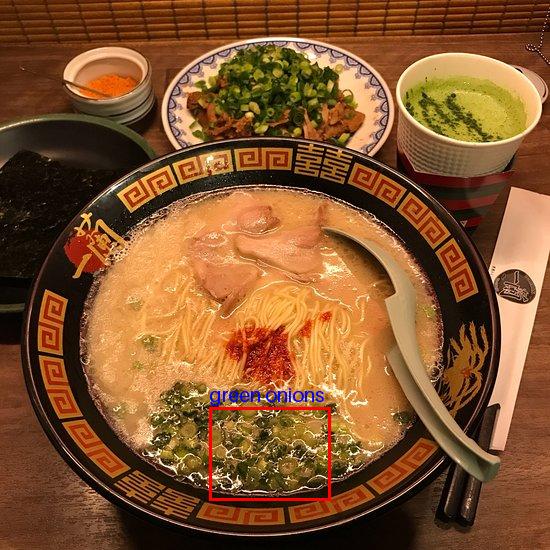}}
      \caption{AIMv2}
      \label{fig:subd}
    \end{subfigure}
    \caption{Output for prompt: \textit{Locate the pieces of green onions in the ramen bowl. Provide the bounding box coordinates given the size of the image is 550x550.}}
    \label{fig:group2}
  \end{subfigure}

  \caption{Object localization in LLaVA-SigLIP2 vs. LLaVA-AIMv2 for different prompts.}
  \label{fig:bounding_box_siglip2_aimv2}
\end{figure}

Table~\ref{tab:cvbench-2d} shows that Qwen2.5-VL achieves the strongest results among frontier models on all CV-Bench 2D tasks. Among our LLaVA variants, LLaVA-AIMv2 excels in CV-Bench 2D subtasks such as COCO and ADE20K. This is because the model’s pre-training objective aligns with the nature of these tasks. COCO is for object detection and instance segmentation, and ADE20K is a scene parsing benchmark. Performance on these subtasks of CV-Bench 2D depends on the model’s ability to capture fine-grained, pixel-level representations of spatial relationships. On the other hand, CLIP and SigLIP perform less effectively on these tasks. Their core objective is to learn a global vector for image-text matching, not a dense feature map. This limitation prevents them from achieving fine-grained visual understanding. Consequently, adding 2D-RoPE only helps to organize these global-level representations from these encoders and does not enable the model to learn any dense representations. SigLIP2’s caption-based pretraining from LocCa \citep{wan2024loccavisualpretraininglocationaware} along with self-distillation and masked prediction from SILC \citep{naeem2023silcimprovingvisionlanguage} and TIPS \citep{maninis2025tipstextimagepretrainingspatial} explicitly directs the model to associate language to specific and relevant regions of the image. As a result, we see that SigLIP2 performs better than its predecessor in the subtasks of CV-Bench 2D. 

An interesting observation in Table~\ref{tab:cvbench-2d} is that LLaVA-SigLIP2 lags behind LLaVA-AIMv2 which might be unexpected given SigLIP2's masked image prediction objective. We think this is because the dense supervision is a secondary objective in SigLIP2. Dense supervision losses are added during the last 20\% of the training in SigLIP2\citep{tschannen2025siglip2multilingualvisionlanguage}. On the other hand, the core training objective of AIMv2 is based on learning the dense features from the very beginning. A qualitative example is shown in Figure~\ref{fig:bounding_box_siglip2_aimv2} to demonstrate how LLaVA-AIMv2 is more precise in locating objects than LLaVA-SigLIP2.

As shown in Table~\ref{tab:gqa-breakdown}, LLaVA-NeXT leads the GQA Query and Overall performance. PaliGemma leads GQA Choose and Logical. Among our trained LLaVA variants, the SigLIP-based models stand out: LLaVA-SigLIP-2D-RoPE achieves the highest scores on Overall, Choose, Compare, and Query, while LLaVA-SigLIP2-2D-RoPE leads on Logical and Verify. LLaVA-AIMv2-2D-RoPE shows moderate improvements over its baseline, particularly on Compare and Logical, but it does not surpass the SigLIP-based variants. We observe that Gemma3-4b-it surpasses the LLaVA variants, including the 2D-RoPE variants of SigLIP and SigLIP2 for GQA Choose subtask. We think this is because how Gemma3 is trained on supervised distillation from a large teacher model with a frozen visual encoder, making it an effective classifier. The Choose subtask of GQA is a constrained multi-task classification problem that aligns well with the training benefits of Gemma3. For other GQA subtasks, LLaVA 2D-RoPE variants of SigLIP, SigLIP2, and AIMv2 surpass Gemma3. We think that the structural advantage of 2D-RoPE performs well with the object and attribute relationships like GQA Compare, Logical. For substasks like GQA Query and Verify, the model needs to confirm information related to a specific object or attribute. SigLIP2’s self-supervised losses and masked prediction result in rich local representations that are needed for these complex, compositional reasoning.

\begin{table*}[h]
    \centering
    \caption{Comparison of frontier models and LLaVA variants on CV-Bench 2D tasks. Values \underline{underlined} indicate the best-performing model overall, while values in \textbf{bold} highlight the best-performing LLaVA encoder variant in each benchmark.}
    \begin{tabular}{lccc}
        \toprule
        Models & \makecell{CV-Bench \\ 2D COCO} & \makecell{CV-Bench \\ 2D ADE20K} & \makecell{CV-Bench \\ 2D Overall} \\
        \toprule
        LLaVA-NeXT & 0.680 & 0.532 & 0.606 \\
        \nobrk{LLaVA-OneVision} & 0.775 & 0.684 & 0.730 \\
        Qwen2.5-VL & \underline{0.820} & \underline{0.689} & \underline{0.754} \\
        SmolVLM2 & 0.621 & 0.532 & 0.577 \\
        \nobrk{Gemma3-4b-it} & 0.660 & 0.618 & 0.659 \\
        PaliGemma & 0.675 & 0.573 & 0.624 \\
        Molmo & 0.773 & 0.684 & 0.728 \\
        LLaVA-1.5 & 0.525 & 0.455 & 0.490 \\
        \midrule
        \nobrk{LLaVA-2D-RoPE} & 0.461 & 0.425 & 0.443 \\
        LLaVA-SigLIP & 0.440 & 0.384 & 0.412 \\
        \nobrk{LLaVA-SigLIP-2D-RoPE} & 0.445 & 0.404 & 0.425 \\
        LLaVA-SigLIP2 & 0.466 & 0.419 & 0.442 \\
        \nobrk{LLaVA-SigLIP2-2D-RoPE} & 0.436 & 0.393 & 0.415 \\
        LLaVA-AIMv2 & \textbf{0.480} & \textbf{0.453} & \textbf{0.466} \\
        \nobrk{LLaVA-AIMv2-2D-RoPE} & 0.456 & 0.408 & 0.432 \\
        \bottomrule
    \end{tabular}
    \label{tab:cvbench-2d}
\end{table*}

\begin{table*}[h]
    \centering
    \caption{Comparison of frontier models and LLaVA variants on GQA subtasks. Values \underline{underlined} indicate the best-performing frontier model; values in \textbf{bold} indicate the best-performing LLaVA variant.}
    \begin{tabular}{lcccccc}
        \toprule
        Models & \makecell{GQA \\ Overall} & \makecell{GQA \\ Choose} & \makecell{GQA \\ Compare} & \makecell{GQA \\ Logical} & \makecell{GQA \\ Query} & \makecell{GQA \\ Verify} \\
        \toprule
        LLaVA-NeXT & \underline{63.786} & 85.120 & 64.177 & 78.980 & \underline{49.875} & 82.860 \\
        \nobrk{LLaVA-OneVision} & 62.140 & 83.880 & 66.893 & 80.588 & 46.069 & \underline{83.792} \\
        Qwen2.5-VL & 60.391 & 84.322 & \underline{73.854} & 78.702 & 43.027 & 82.682 \\
        SmolVLM2 & 50.574 & 67.434 & 47.538 & 64.504 & 37.867 & 70.160 \\
        \nobrk{Gemma3-4b-it} & 31.277 & 61.913 & 16.978 & 19.301 & 25.952 & 45.337 \\
        PaliGemma & 62.570 & \underline{86.802} & 73.345 & \underline{81.864} & 45.893 & 82.549 \\
        Molmo & 55.295 & 78.742 & 52.462 & 65.169 & 41.558 & 77.886 \\
        LLaVA-1.5 & 33.225 & 49.690 & 49.576 & 33.500 & 25.702 & 43.206 \\
        \midrule
        \nobrk{LLaVA-2D-RoPE} & 34.433 & 41.807 & 54.839 & 48.697 & 22.337 & 50.533 \\
        LLaVA-SigLIP & 25.648 & 17.006 & 47.199 & 44.204 & 15.783 & 39.298 \\
        \nobrk{LLaVA-SigLIP-2D-RoPE} & \textbf{38.448} & \textbf{55.182} & \textbf{59.762} & 55.186 & \textbf{23.159} & 57.282 \\
        LLaVA-SigLIP2 & 23.970 & 16.475 & 40.917 & 41.486 & 14.107 & 39.076 \\
        \nobrk{LLaVA-SigLIP2-2D-RoPE} & 34.560 & 47.121 & 55.688 & \textbf{57.959} & 15.298 & \textbf{62.211} \\
        LLaVA-AIMv2 & 32.541 & 37.998 & 51.783 & 47.421 & 21.440 & 46.403 \\
        \nobrk{LLaVA-AIMv2-2D-RoPE} & 32.342 & 39.858 & 56.876 & 48.974 & 18.663 & 50.178 \\
        \bottomrule
    \end{tabular}
    \label{tab:gqa-breakdown}
\end{table*}

\section{Conclusion}
\label{sec:conclusion}
We demonstrate that encoder choice significantly impacts spatial reasoning. For example, CountBenchQA improves about 58\%, rising from 0.468 in LLaVA-1.5 to 0.739 in LLaVA-AIMv2. The effects of injecting 2D-RoPE into the image encoder attention are mixed, indicating that where and how 2D positional information is introduced matter. In our experiments, frontier models such as Qwen2.5-VL achieve the strongest overall results. Although Qwen2.5-VL is trained on a different dataset and token scales than LLaVA, the comparison is not strictly apples-to-apples.  Overall, the findings highlight that encoder design shapes spatial awareness within VLM families, although comparisons to frontier models remain questionable due to differences in training data and scale.
\section{Future Work}
\label{sec:futurework}

Our study focused on static, 2D images, benchmarks and encoder variants within the LLaVA framework. This work can extend to 3D spatial reasoning along with the dynamic environment. Another potential extension can be on SigLIP2 with NaFlex. The flexible resolution image preprocessing of NaFlex mitigates information loss observed in fixed-resolution encoders. Similarly BLIP-2 and related architectures could help assess whether their pretraining objectives and visual-language alignment strategies offer advantages over CLIP-derived models. Finally, we note that advanced alignment mechanisms, such as gated attention in Flamingo \cite{alayrac2022flamingo}, Q-Former in BLIP-2, or cross-modal pooling in MM1 \cite{mckinzie2024mm1}, were intentionally left out in this work. Exploring these approaches as alternatives to simple projection layers may further improve spatial reasoning performance.
\section{Acknowledgment}
\label{sec:ack}

We gratefully acknowledge EleutherAI for providing GPU resources that supported this work. We also thank the volunteer team members for their time, dedication, and invaluable contributions.
{
    \small
    \bibliographystyle{ieeenat_fullname}
    \bibliography{main}
}


\end{document}